\documentclass[10pt,twocolumn,letterpaper]{article}

\usepackage[pagenumbers]{cvpr} 

\usepackage[dvipsnames]{xcolor}

\definecolor{cvprblue}{rgb}{0.21,0.49,0.74}
\usepackage[pagebackref,breaklinks,colorlinks,citecolor=cvprblue]{hyperref}

\usepackage{algorithm}
\usepackage{algpseudocode}

\def\paperID{} 
\def\confName{3DV\xspace}
\def\confYear{2026\xspace}

\title{MVRoom: Controllable 3D Indoor Scene Generation with Multi-View Diffusion Models}

\author{
    Shaoheng Fang$^{1}$\thanks{Work done during internship at DAMO Academy, Alibaba Group} \quad
    Chaohui Yu$^{2,3}$ \quad
    Fan Wang$^2$ \quad
    Qixing Huang$^1$ \\[5pt]
    $^1$The University of Texas at Austin, $^2$DAMO Academy, Alibaba Group, $^3$Hupan Lab
}

\begin{document}
\maketitle
\begin{abstract}
We introduce MVRoom, a controllable novel view synthesis (NVS) pipeline for 3D indoor scenes that uses multi-view diffusion conditioned on a coarse 3D layout. MVRoom employs a two-stage design in which the 3D layout is used throughout to enforce multi-view consistency. The first stage employs novel representations to effectively bridge the 3D layout and consistent image-based condition signals for multi-view generation. The second stage performs image-conditioned multi-view generation, incorporating a layout-aware epipolar attention mechanism to enhance multi-view consistency during the diffusion process. 
Additionally, we introduce an iterative framework that generates 3D scenes with varying numbers of objects and scene complexities by recursively performing multi-view generation (MVRoom), supporting text-to-scene generation. Experimental results demonstrate that our approach achieves high-fidelity and controllable 3D scene generation for NVS, outperforming state-of-the-art baseline methods both quantitatively and qualitatively. Ablation studies further validate the effectiveness of key components within our generation pipeline.
\end{abstract}    


\section{Introduction}
\label{sec:intro}

\begin{figure*}[t]
\centering
\includegraphics[width=0.95\linewidth]{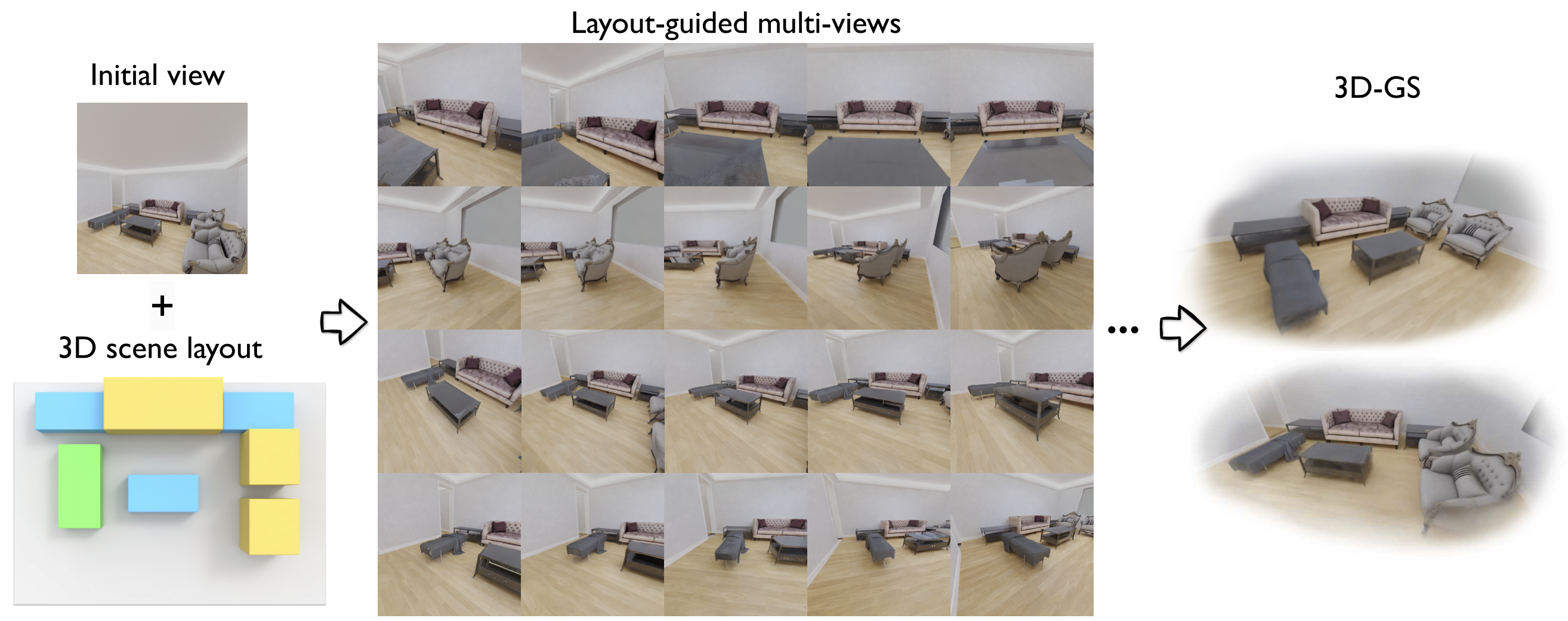}
\vspace{-3mm}
\caption{We introduce MVRoom, an indoor scene generation pipeline utilizing multi-view diffusion models. Given a 3D layout and an initial image (generated from a text description). MVRoom uses conditional layout-aware multi-view diffusion models to generate consistent novel views along continuous camera trajectories within the 3D scene. The consistent views are fed into a 3D-GS pipeline for scene reconstruction and novel-view synthesis. }
\vspace{-3mm}
\label{fig:fig1_overview}
\end{figure*}


Creating high-quality 3D content is crucial for immersive applications in augmented reality (AR), virtual reality (VR), gaming, and filmmaking. Yet producing detailed 3D assets remains challenging and labor-intensive, often requiring professional skills and tools. Recent advances in generative modeling enable text-to-3D object synthesis, streamlining creation~\cite{poole2022dreamfusion, lin2023magic3d}. However, generating 3D scenes is more complex, as it involves creating numerous objects arranged within complex spatial layouts, and the generated scenes must exhibit realism to ensure authenticity and user engagement.


Several methods have recently been proposed to address 3D scene generation~\cite{cohen2023set, zhou2024gala3d, zhang2023scenewiz3d, zhang2024text2nerf, hollein2023text2room, ouyang2023text2immersion, ma2024fastscene, schult2024controlroom3d, fang2023ctrl}. Composition approaches first generate objects via text-to-3D techniques and then assemble a full scene~\cite{cohen2023set, zhou2024gala3d, zhang2023scenewiz3d}. While these methods leverage the strengths of 3D object generation, they frequently fall short in overall scene realism. 
Other methods adopt incremental generation frameworks. They construct 3D indoor environments by sequentially synthesizing different viewpoints frame by frame and reconstructing room meshes from these images \cite{zhang2024text2nerf, hollein2023text2room, ouyang2023text2immersion}. 
However, these methods always suffer from error accumulation and weak layout control, making it difficult to ensure a coherent spatial arrangement.

To address these limitations, we propose a novel approach MVRoom for multi-view-based 3D scene generation, conditioned on initial scene image(s) and a coarse 3D layout (oriented object bounding boxes with class labels). The 3D layout, which can be obtained from user input or off-the-shelf 3D layout generative models, provides a flexible and easily editable representation, serving as essential guidance of our framework. Moreover, the initial images can be generated using text-2-image models (see Figure~\ref{fig:fig7_text_to_scene} and the accompanying video).

MVRoom has two stages. The first stage focuses on how to bridge the 3D layout into image-based condition signals for multi-view generation. To achieve this, we introduce a novel image-based representation that preserves 3D layout information and enables cross-view feature aggregation during generation. This representation includes hybrid layout priors, incorporating multi-layer semantics and depth conditions, along with layout spatial embeddings. In the second stage, we employ an image-conditioned multi-view generative model that produces views that cover the underlying scene, supporting the whole scene generation. In addition to conditioning on rich signals derived from a 3D layout, we introduce a layout-aware epipolar attention layer to effectively fuse cross-view features. This approach ensures accurate feature alignment across perspectives, greatly enhancing multi-view consistency in the generated images, a critical issue in multi-view based 3D generation.

In addition, to generate a complete indoor scene, we adopt an iterative framework that recursively produces multi-view content according to the 3D scene layout. Our framework takes as input the 3D layout and one or more initial scene images—potentially synthesized from text—to guide the style of the generated scene. To handle varying object counts and spatial ranges, we first automatically generate camera trajectories that explore different parts of the layout, then perform multi-view generation along each trajectory. 
To ensure global consistency, we maintain a global point cloud and render images from it as the image conditions for new generations, ensuring cross-view consistency. We initialize and selectively update the global point cloud using layout-aware monocular depth predictions.

We evaluate MVRoom on the 3D-Front dataset~\cite{3dfront}, generating multi-view data for training and testing. Our experiments show that the proposed image-based representations, designed to encode 3D layout priors can effectively guide the generation of realistic and coherent 3D scenes. We also present an ablation study that demonstrates the effectiveness of key components in our pipeline, including the image-based representations and the layout-aware epipolar attention mechanism for multi-view generation.
Moreover, our results indicate that MVRoom produces 3D scenes that are more globally coherent than those from existing methods. Finally, by optimizing a 3D Gaussian-splatting neural representation~\cite{kerbl20233d} with our generated multi-view data, we can reconstruct a complete, high-fidelity indoor scene and render high-quality content from arbitrary viewpoints.

\noindent In summary, our contributions are as follows:
\begin{itemize}
\item We propose MVRoom, a two-stage method for high-fidelity and controllable novel view synthesis conditioned on a 3D scene layout.
\item We introduce a novel image-based representation under given camera poses to preserve as much information from the input 3D layout as possible.
\item We develop a layout-aware epipolar attention layer within a multi-view diffusion model, ensuring robust cross-view consistency and accurate feature alignment.
\item We design an recursive generation framework with MVRoom, supporting text-to-scene generation. The framework systematically covers the entire scene by sampling camera trajectories according to the 3D layout, while maintaining a global point cloud to ensure consistent generations.
\end{itemize}

\section{Related Works}
\label{sec:related_work}

\subsection{3D Object Generation}

\begin{figure*}[t]
\centering
\includegraphics[width=0.95\linewidth]{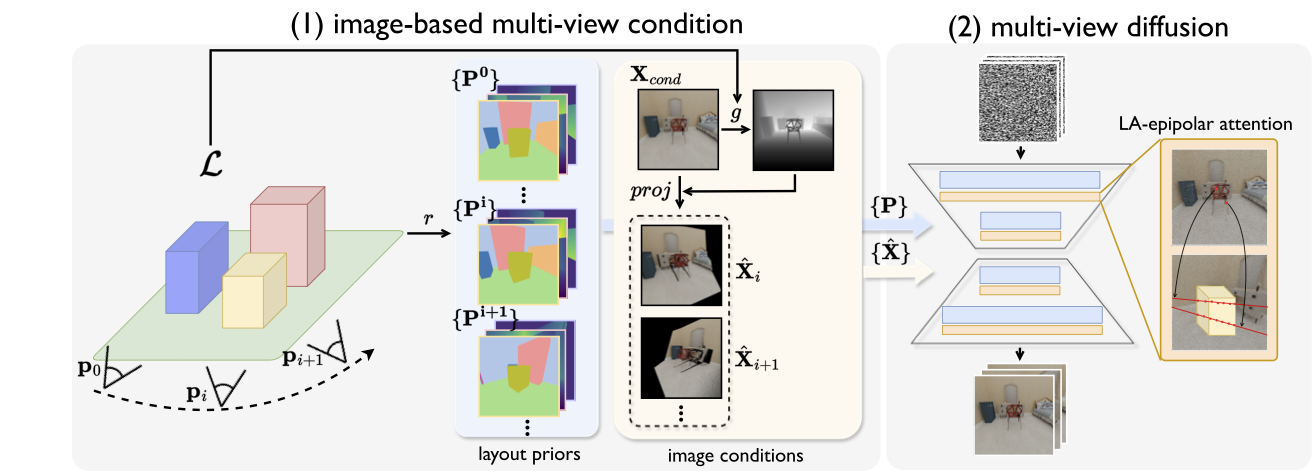}
\vspace{-3mm}
\caption{MVRoom overview. We employ a two-stage generation pipeline. The first stage focuses on gathering image-based multi-view conditions, including hybrid layout priors derived from 3D scene layout $\mathcal{L}$ and camera poses $\mathbf{p}_i$ and image conditions $\mathbf{P}^i$ derived from initial views or generated views. The second stage is a multi-view diffusion model that takes the image-based conditions as input and generates scene-level consistent views. The diffusion model features a layout-aware epipolar attention module to aggregate cross-view features more efficiently according to epipolar geometry and 3D layout.}
\vspace{-3mm}
\label{fig:fig3_architecture}
\end{figure*}

With the rapid advancements in text-to-image models, recent research has focused on leveraging these pre-trained 2D models to guide 3D scene generation. \cite{mohammad2022clip, jain2022zero} use the pre-trained CLIP model~\cite{radford2021learning} to guide mesh or NeRF optimizations. Since DreamFusion~\cite{poole2022dreamfusion}, 2D diffusion priors have become widely used in 3D object generation. DreamFusion~\cite{poole2022dreamfusion} introduced score distillation sampling (SDS) to optimize 3D object creation from text prompts. Subsequent studies have sought to improve the quality of generative processes by exploring diverse 3D representations~\cite{chen2023fantasia3d, lin2023magic3d, tsalicoglou2024textmesh, wu2024hd}, refining distillation strategies~\cite{chen2024it3d, huang2023dreamtime, seo2023ditto}, and designing optimization loss functions~\cite{wang2024prolificdreamer, zhu2023hifa}. Addressing limited 3D consistency with 2D diffusion models alone, \cite{qian2023magic123, weng2023consistent123} propose learning 3D representations via a blend of 2D diffusion priors and view-dependent 3D diffusion cues~\cite{liu2023zero}. However, these approaches do not apply for 3D scene generation, as 1) we do not have the notion of absolute image poses, 2) we need a varying number of images to cover a 3D scene, and 3) it is difficult to enforce multi-view consistency.


\subsection{Image-based Novel View Synthesis}



Recently, 2D diffusion models have been widely adopted for synthesizing novel 3D views from single images.
\cite{watson2022novel, liu2023zero, liu2023syncdreamer} aim to produce novel 2D view images conditioned on an input image and camera pose. Notably, Zero-1-to-3~\cite{liu2023zero}, fine-tuned directly from robust Stable Diffusion checkpoints on a large 3D object dataset~\cite{deitke2023objaverse}, offers a high-quality 3D diffusion prior to object generation~\cite{qian2023magic123, weng2023consistent123}. 
MVDream~\cite{shi2023mvdream} extends the 2D diffusion model by integrating a 3D self-attention layer, simultaneously denoising four views of an object to ensure consistency. 
To better incorporate 3D priors from pose conditions into the denoising process, MVdiffusion~\cite{Tang2023mvdiffusion} establishes dense alignment across views by leveraging ground-truth depth and homography transformation, employing correspondence-aware attention to aggregate features among multiple perspectives.
Meanwhile, \cite{yu2024viewcrafter, muller2024multidiff} utilize video diffusion prior for 3D scene novel view synthesis.
\cite{zhou2023sparsefusion, kant2024spad, tseng2023consistent} use epipolar transformers~\cite{he2020epipolar} to merge features from input views.
Our approach is inspired by these approaches, but focuses on scene-level image-based novel view synthesis controlled by the 3D scene layout.

\subsection{3D Scene Generation}

Recent advances now enable the generation of 3D scenes directly from textual descriptions. \cite{hollein2023text2room, zhang2024text2nerf, ouyang2023text2immersion, schult2024controlroom3d, fang2023ctrl, fridman2024scenescape, shriram2024realmdreamer, ma2024fastscene} utilize text-to-image methods, iteratively expanding the content of a 3D scene by inpainting models. Depth estimation and alignment methods are utilized to project generated images into the 3D space, progressively merging new content to construct a complete 3D scene. Although this approach can produce photo-realistic results, it often suffers from accumulated errors during the iterative inpainting process, making it challenging to maintain geometric and textural consistency. To mitigate this limitation, \cite{Tang2023mvdiffusion, mao2023showroom3d, song2023roomdreamer, nguyen2024housecrafter} adopt multi-view generation strategies to improve scene coherence. 
Our approach follows this idea and further enhances multi-view consistency by proposing a layout-aware epipolar attention module without the need of ground truth depth maps~\cite{Tang2023mvdiffusion, nguyen2024housecrafter}. 


%

Another line of research employs text-to-3D object generation techniques~\cite{poole2022dreamfusion, shi2023mvdream}, constructing 3D scenes by generating individual 3D objects and integrating them according to a predefined 3D scene layout~\cite{cohen2023set, bai2023componerf, li2024dreamscene, zhang2023scenewiz3d, zhou2024gala3d}. These methods benefit from the strong priors provided by 3D scene layouts, facilitating controllable and spatially coherent results. However, they generally produce scenes of incompatible objects because of the independent generation of various objects.
Some approaches have incorporated scene layouts into holistic scene generation frameworks to balance quality and controllability~\cite{schult2024controlroom3d, fang2023ctrl, bahmani2023cc3d, nguyen2024housecrafter}. \cite{schult2024controlroom3d, fang2023ctrl} use 3D layouts as guidance for generating panoramas, serving as an initial step in scene generation, but are primarily limited to simple, forward-facing scenes. Likewise, \cite{bahmani2023cc3d, zhou2024gala3d} employ 2D floorplans as references for the generation of indoor scenes, enhancing controllability but with limitations in the capture of 3D spatial relationships between scene elements. 
Scenecraft~\cite{yang2024scenecraft} finetunes a layout-conditioned text-to-image diffusion model to optimize the neural representation~\cite{mildenhall2021nerf} of the entire scene via score distillation sampling (SDS). However, the optimization views are dependent on and constrained to a pre-defined camera trajectory, and the usage of SDS compromises the photorealism of generated results.
Our approach employs hybrid 3D priors from scene layouts, substantially improving 3D consistency and leveraging the comprehensive benefits of 3D layout guidance.

\section{Method}
\label{sec:method}

\begin{figure*}[h]
    \centering
    \includegraphics[width=0.95\linewidth]{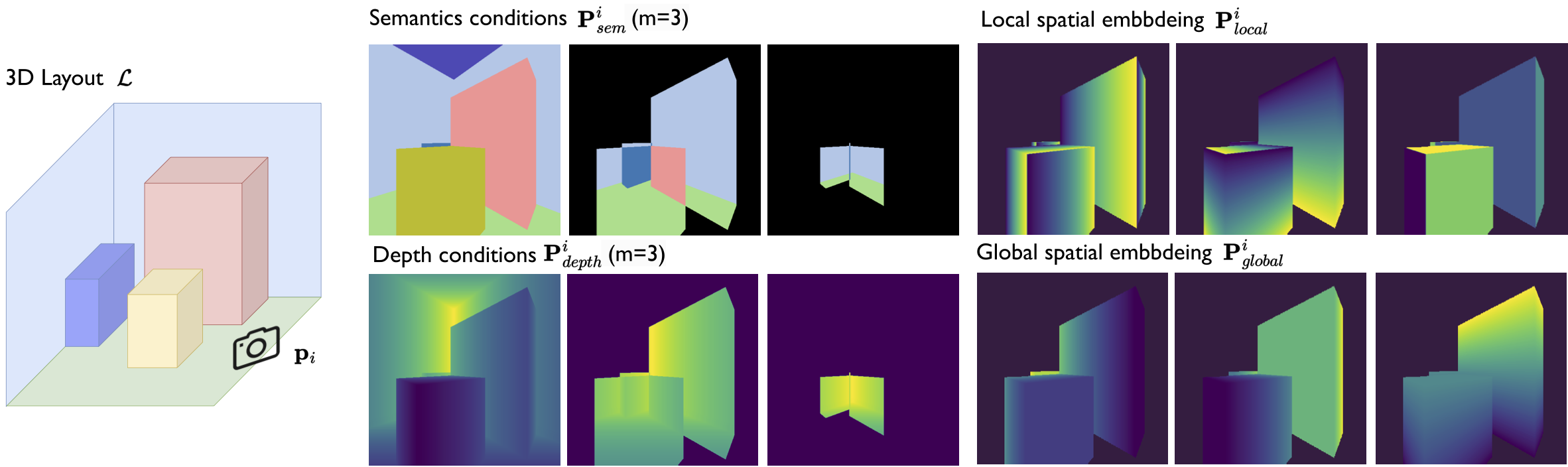}
    \vspace{-3mm}
    \caption{Comprehensive 3D layout conditions. Given a camera parameter $\mathbf{p}_i$, we convert the layout $\mathcal{L}$ into multiple image conditions $\mathbf{P}^i$.}
    \vspace{-3mm}
    \label{fig:fig2_priors}
\end{figure*}


MVRoom is a two-stage novel-view synthesis framework (Figure~\ref{fig:fig3_architecture}). 
The first stage (Section~\ref{sec:hybrid_3d_layout}) focuses on encoding the 3D layout prior and condition image(s) into multi-view-based condition signals. The second stage (Section~\ref{sec:mv_generation}) performs multi-view generation with a novel layout-aware epipolar attention mechanism to enhance cross-view consistency.
Section~\ref{sec:scene_generation} introduces our novel scene generation framework that applies MVRoom recursively to generate complete indoor scenes.

\subsection{Condition Signal Representations}
\label{sec:hybrid_3d_layout}


Condition signals combine hybrid 3D layout priors and image-based priors. For each type, we first describe its original representation. We then introduce its image-based representation that is fed into the condition multi-view generation module in Section~\ref{sec:mv_generation}. 

\noindent\textbf{3D layout condition.} The first condition signal is a 3D scene layout $\mathcal{L}$ that consists of 1) bounding boxes $\mathcal{B} = \{ B_1, \cdots, B_{N_o} \}$, where $N_o$ is the number of objects in the scene, and 2) the background contents, including floors, walls, and ceilings, etc. 
Each bounding box is represented as $B_j = [c_j, \mathbf{s}_j, \mathbf{l}_j, r_j]$, where $c_j \in \{ 1, \cdots, N_c \}$ is the semantic category, $\mathbf{s}_j \in \mathbb{R}^3$ is its size, $\mathbf{l}_j \in \mathbb{R}^3$ is its location and $r_j$ is its rotation angle around the $\mathbf{s}$ axis. 

Given camera $\mathbf{p}_i$, we convert the layout $\mathcal{L}$ into a multi-channel image $\mathbf{P}^i = (\mathbf{P}^i_{\textup{sem}},\mathbf{P}^i_{\textup{depth}},\mathbf{P}^i_{\textup{local}},\mathbf{P}^i_{\textup{global}})$. This image-based encoding is designed to preserve information in the 3D layout as much as possible and facilitate cross-view feature aggregation. 

The semantics and depth channels are given by
\begin{equation}
\label{eq:dns_condition}
    (\mathbf{P}^i_{\textup{sem}}, \mathbf{P}^i_{\textup{depth}}) = r_{\{\textup{sem}, \textup{depth}\}} (\mathcal{L}, \mathbf{p}_i),
\end{equation}
where $r$ is the function that renders the semantic and depth prior map. A novelty of our approach is that we define $\mathbf{P}^i_{\textup{sem}}, \mathbf{P}^i_{\textup{depth}} \in \mathbb{R}^{H\times W \times m}$ to be multi-layer conditions, where $m>1$ is the layer number. As shown in Figure~\ref{fig:fig2_priors}, our approach considers the first $m$ intersection points along each pixel ray with the 3D scene layout $\mathcal{L}$. Additionally, if a ray intersects with the floor or walls, further intersections are ignored, and the subsequent values in the conditions are zero. Compared to only using the first intersection point, our approach captures more comprehensive 3D scene information by encoding occluded geometric signals. 

The spatial embeddings of the layout $\mathbf{P}^i_{\textup{local}}$ and $\mathbf{P}^i_{\textup{global}}$ encode the orientations and absolute locations of the bounding boxes. They are given by
\begin{equation}
    (\mathbf{P}^i_{\textup{local}}, \mathbf{P}^i_{\textup{global}}) = r_{\textup{spatial}} (\mathcal{L}, \mathbf{p}_i).
\end{equation}
Specifically, the local spatial embedding map $\mathbf{P}^i_{\textup{local}} \in \mathbb{R}^{H\times W \times 3}$ describes spatial information at the object level from the perspective view $\mathbf{p}_i$. For each bounding box, we construct a 2D mesh grid on each of its visible surfaces, ranging from $-1$ to $1$ with a resolution of $(h^{\prime}, w^{\prime})$.
The 2D mesh grids on these surfaces are then projected onto the perspective view by perspective transformation, resulting in the first two channels of $\mathbf{P}^i_{\textup{local}}$. The final channel represents the surface index ($1$ to $6$) of the bounding box to which each pixel belongs, determined by the orientation of the bounding box.

The global spatial embedding map $\mathbf{P}^i_{\textup{global}} \in \mathbb{R}^{H\times W \times 3}$ captures the absolute positional information of the bounding box layout $\mathcal{B}$ with the camera view space. This encoding allows the network to easily identify the corresponding regions between two views. Similar to the calculation for local embedding, we first create a 2D grid map for each surface of the bounding box. In this case, the grid map has three channels, where each grid point represents the 3D coordinates of its center in the entire 3D scene. The visible portions of all the surfaces of the boundary box are then projected onto the perspective view $\mathbf{p}_i$, resulting in $\mathbf{P}^i_{\textup{global}}$.

\noindent\textbf{Image condition.} 
In addition to the condition signals from the 3D layout, our approach takes a single scene image $\mathbf{X}_{\textup{cond}}$ as input for novel view synthesis. This condition image, which may be provided by users or generated via a text-to-image model conditioned on $(\mathbf{P}^i_{\textup{sem}}, \mathbf{P}^i_{\textup{depth}})$, establishes the overall style and content of the generated views.
To incorporate additional 3D priors, we adapt $\mathbf{X}_{\textup{cond}}$ to each target view using a predicted depth. During MVRoom’s training and testing, this is achieved by warping the condition image to the target camera pose $\mathbf{p}_i$. During inference, the warping is driven by the predicted depth map $\mathbf{\hat{D}}_{\textup{cond}}$ of $\mathbf{X}_{\textup{cond}}$. We initialize $\mathbf{D}_{\textup{cond}}$ using an off-the-shelf depth estimation method~\cite{yang2024depth} to predict the metric depth of the condition image. Since Eq.~(\ref{eq:dns_condition}) provides a coarse approximation of the underlying 3D structure, we further rectify the estimated depth using the 3D layout, denoting as $\mathbf{\hat{D}}_{\textup{cond}} = g(\mathbf{X}_{\textup{cond}}, \mathcal{L})$.
Please refer to the supp. materials for the details.  

With the rectified depth $\mathbf{\hat{D}}_{\textup{cond}}$, we utilize \cite{niklaus2020softmax} to warp the condition image to each view: 
\begin{equation}
\label{eq:warp_img}
    \mathbf{\hat{X}}_i = proj(\mathbf{X}_{\textup{cond}}, \mathbf{p}_{\textup{cond}}, \mathbf{p}_{i}, \mathbf{\hat{D}}_{\textup{cond}}).
\end{equation}
Finally, we augment $\mathbf{\hat{X}}_i$ with its inpainting mask as the final image condition of view $\mathbf{p}_i$.

\subsection{Condition Multi-view Generation}
\label{sec:mv_generation}


In the following, we detail the multi-view diffusion model and explain how the layout and image conditions are injected. We also emphasize our novel layout-aware epipolar attention design, which is crucial for robust cross-view consistency.

Our model primarily employs a multi-view diffusion model that generates $N$ views of the underlying scene $\{ \mathbf{X}_i \in \mathbb{R}^{H\times W \times 3} \}_{i=1}^N$ conditioned on camera poses $\{ \mathbf{p}_i \}_{i=1}^N$ and hybrid priors $\{ \mathbf{P}^i \}_{i=1}^N$ and the image conditions $\hat{\mathbf{X}}_{i}$ described in the preceding section. Formally speaking, this module is modeled as: 
\begin{equation}
    p(\{ \mathbf{X}_i \}_{i=1}^N | \{\hat{\mathbf{X}}_{i}\}_{i=1}^N, \{ \mathbf{P}^i \}_{i=1}^N, \{ \mathbf{p}_i \}_{i=1}^N).
\end{equation}

The overall architecture is illustrated in Figure~\ref{fig:fig3_architecture}.
It is based on a text-to-image latent diffusion model (LDM), initialized from a pre-trained text-to-image model using web-scale datasets~\cite{rombach2022high}.
Hybrid 3D layout priors are incorporated into the generative process via separate adapters~\cite{mou2024t2i} attached to the LDM, allowing the model to effectively handle complex 3D scene information.

Central to our approach is a layout-based epipolar attention mechanism. The goal is to ensure effective cross-view feature integration, which is critical yet challenging for multi-view generation of complex 3D scenes with arbitrary camera poses. Previous scene-level multi-view generation methods~\cite{Tang2023mvdiffusion, nguyen2024housecrafter} establish dense pixel correspondences according to relative homography transformations. However, they rely on ground-truth depth prior and struggle to handle occlusion issues arising from viewpoint changes. There are recent approaches that leverage the epipolar attention mechanism~\cite{tseng2023consistent,DBLP:conf/eccv/WangXTCLPFKZ24}. The difference in our formulation is that we explicitly integrate 3D layout information into the attention module. 

Specifically, our attention mechanism, denoted as $f_{\rm EpiAttn}$, leverages both epipolar geometry and 3D scene layout to select feature maps in cross-view attention.
Denote $\mathbf{F}^i$ as the feature map of the $i$-th image, and denote $\mathbf{F}^r$ as the feature map of the remaining $N-1$ views. 
$\tilde{\mathbf{F}}^r = \mathcal{M}^{i, \mathcal{R}}_{\rm LA}(\mathbf{F}^r)$ is the masked feature based both on the epipolar geometry and on the 3D layout, where $\mathcal{M}^{i, \mathcal{R}}_{\rm LA}$ is the mask function. For a pixel $q$ in $\mathbf{F}_i$, the layout-ware epipolar attention can be written as:
\begin{equation}
\label{eq:epipolar}
    \begin{split}
    f_{\rm EpiAttn} (\mathbf{F}^i, \mathbf{F}^r, q) = & \\
    {\rm SoftMax}(Q(\mathbf{F}^i_{q}) & \cdot K([\mathbf{F}^i | \tilde{\mathbf{F}}^r])) V([\mathbf{F}^i | \tilde{\mathbf{F}}^r)]),
    \end{split}
\end{equation}
where $Q, K, V$ are the query, key, and value models and $[|]$ is the feature concatenation operation. 



We proceed to define the mask function $\mathcal{M}^{i, \mathcal{R}}_{\rm LA}$.
For the pixel $q$ in $\mathbf{F}^i$, we can sample any positive depth value $d$ of $q$ and calculate the corresponding 3D points ${q}_j(d)$ in the $j$-th image's camera coordinate system and its projected image coordinate system $q_j(d)$. The epipolar line is given by $q_j(d)$ with varying $d$. 
In a 3D scene with known 3D layouts $\mathcal{L}$, obviously only a small segment of the ray from $q$ overlaps with the layout, making only this part of the features in the epipolar line relevant and effective for the aggregation of features of
$\mathbf{F}^i_{q}$. In our implementation, we calculate all depths at which the ray from $q$ intersects the layout $\mathcal{L}$. Denote $d^k_1$ and $d^k_2$ as the two depths where the ray intersects with a bounding box $B_k$, only the epipolar line between $q_j(d^k_1)$ and $q_j(d^k_2)$ is the portion we need. 


\subsection{Recursive Scene Generation}
\label{sec:scene_generation}

\begin{algorithm} 
\vspace{-3mm}
\caption{Recursive Scene Generation (Text-to-Scene)}\label{alg:cap}
\begin{flushleft}
\textbf{Input:} An initial view $\mathbf{p}_0$, 3D layout $\mathcal{L}$, text prompt $text$, layout-guided text-to-image model $SD\_adapter(\cdot)$, depth estimation model $g(\cdot)$, layout-guided novel view synthesis model $MVRoom(\cdot)$.
\end{flushleft}
\begin{algorithmic}[1]
\State \textbf{Initialize:} $PC_{\text{global}} \leftarrow \emptyset$, $\text{gen\_idx} = \{\}$, $\text{gen\_img} = \{\}$
\State $\mathbf{X}_0 \leftarrow SD\_adapter(\mathcal{L}, \mathbf{p}_0, text)$
\State Sample $N_p$ view poses from multiple trajectories starting from $p_0$: $\{ \mathbf{p}_0, \cdots, \mathbf{p}_{N_p} \}$
\State $PC_{\text{cur}} \leftarrow ProjectPC(\mathbf{X}_0, g(\mathbf{X}_0, \mathcal{L}))$
\State $PC_{\text{global}} \leftarrow PC_{\text{global}} \cup PC_{\text{cur}}$

\While{unfinished}
\State Sample $N$ ungenerated consecutive poses from one trajectory $\mathbf{p}_{k_1}, \cdots, \mathbf{p}_{k_N}$
\State Get ref images $\{\hat{\mathbf{X}}_{k_i}\}_{i=1}^N \leftarrow \{ Render(PC_{\text{global}}, \mathbf{p}_{k_i}) \}_{i=1}^N$
\State Get conditions $\{ \mathbf{P}^{k_i} \}_{i=1}^N$  \Comment{Refer to Sec 3.1.}
\State $\{ \mathbf{X}_{k_i} \}_{i=1}^N \leftarrow MVRoom( \{\hat{\mathbf{X}}_{k_i}\}_{i=1}^N, \{ \mathbf{P}^{k_i} \}_{i=1}^N, \{ \mathbf{p}_{k_i} \}_{i=1}^N )$  \Comment{Generate novel views.}
\For{$i = 1, \cdots, N$}
    \State $\text{gen\_idx}.append(k_i) $
    \State $\text{gen\_img}.append(\hat{\mathbf{X}}_{k_i}) $
    \State $\mathbf{\hat{D}}_{k_i} \leftarrow g(\mathbf{X}_{k_i}, \mathcal{L})$
    \State $\mathbf{D}_{k_i} \leftarrow Render\_depth (PC_{\text{global}}, \mathbf{p}_{k_i})$
    \If{$Check\_depth\_consistency ( \mathbf{\hat{D}}_{k_i}, \mathbf{D}_{k_i} )$}
        \State $PC_{\text{cur}} \leftarrow ProjectPC(\mathbf{X}_{k_i}, g(\mathbf{X}_{k_i}, \mathcal{L}))$
        \State $PC_{\text{global}} \leftarrow PC_{\text{global}} \cup PC_{\text{cur}}$
    \EndIf
\EndFor
\EndWhile
\State Optimize 3DGS with generated images
\end{algorithmic}
\label{Algorithm1}
\end{algorithm}
\vspace{-3mm}


While our multi-view generation method can synthesize coherent views given a fixed set of camera poses, it is insufficient to cover a complete indoor environment, especially those with multiple objects and complex spatial configurations. To address this, we extend MVRoom into a recursive generation framework that progressively expands scene coverage while maintaining global consistency and high fidelity. This framework ensures comprehensive scene generation, effectively handling varying object counts, spatial ranges, and layout complexities.  The algorithm~\ref{Algorithm1} provides the pseudocode, using text-to-scene as an example task. 

First, the framework takes one initial image from an arbitrary viewpoint, which can be user-provided or synthesized from text prompts, to define the scene's style and appearance. These images serve as the initial condition for the first batch of multi-view generation, and the content generated from each batch is recursively incorporated as the condition for subsequent batches. This flexible initialization accommodates diverse use cases without altering the core pipeline.

Selecting appropriate camera poses within a complex 3D scene for multi-view generation is a highly non-trivial task. First, all camera views must fully cover the visual content within the scene. Moreover, it is essential to ensure that during multi-view generation, content among adjacent views overlaps to facilitate quality and maintain consistency in generation. To determine reasonable and comprehensive views, we generate a series of continuous camera trajectories according to the 3D scene layout. Each camera trajectory is primarily designated to focus on the bounding box of a specific object, aiming to achieve complete coverage of that object. The trajectory positions are carefully sampled to maintain a reasonable distance from the object while ensuring physical plausibility within the 3D layout.
Additionally, smooth and continuous transitions in both displacement and viewpoint are ensured along the camera trajectory. Please refer to the supp. material for details. 

Another critical challenge in recursive multi-view generation is ensuring consistent visual content across multiple batches. To address this, we maintain a global point cloud throughout the iterative process, inspired by ViewCrafter~\cite{yu2024viewcrafter}. Instead of predicting depth for each condition image and warping it for each new view, we render the point-based image condition from the continually updated global point cloud. This point cloud is initially constructed from the depth of the initial image and is progressively expanded by integrating the predicted depths from newly generated views. Meanwhile, to ensure geometric consistency, we exclude generated results that exhibit noticeable inconsistencies with the current scene geometry before adding them to the global point cloud by comparing the predicted depth of the generated views and the rendered depth from the global point cloud, mitigating potential visual artifacts as new views are incorporated. More details are provided in the supp. material. 

Overall, during generation, MVRoom is applied recursively along each camera trajectory to generate new content for each view. The generated views and their corresponding depths are selectively integrated into the global point cloud, gradually improving the geometric fidelity of the scene. This approach enables the entire indoor environment, which spans multiple objects and spatial regions, to be recursively traversed and generated with maximum consistency. The resulting collection of multi-view images can be used to optimize a 3D Gaussian-splatting neural representation~\cite{kerbl20233d}, enabling high-fidelity reconstruction and free-viewpoint rendering of the complete indoor scene. Please refer to the supp. materials for more
details about the implementation of the recursive scene generation framework.

\section{Experiments}
\label{sec:experiments}

\subsection{Experimental Settings}

\noindent \textbf{Datasets.}
We conduct experiments on the 3D indoor scene dataset 3D-FRONT~\cite{3dfront}, which contains 18,797 professionally designed rooms populated with high-quality, textured 3D models from 3D-FUTURE~\cite{3dfuture}. 3D-FRONT provides extensive indoor scene data, with key advantages over other mainstream indoor scene datasets~\cite{chang2017matterport3d, yeshwanth2023scannet++, roberts2021hypersim} in that we can obtain accurate 3D scene layouts and render flexible multi-view images from arbitrary perspectives.
For MVRoom training and evaluation, we filter a subset of 6,287 rooms from the 3D-FRONT dataset (5,887 for training and 400 for validation and testing).
To generate images from diverse viewpoints within each scene, we first generate multiple continuous random-walk camera trajectories within the room and render images along these paths. To ensure continuity in camera views, camera views along each trajectory are set to point toward a randomly chosen fixed object or the center of the room.
We sample approximately 700 viewpoints per scene, resulting in over 4 million camera views in the entire dataset. RGB data and other conditions are rendered at a resolution of $512 \times 512$.
Moreover, to promote data diversity and improve model robustness, we introduce randomization in rendering settings, including lighting conditions, background colors, and style.
During training, we randomly select $N$ frames from each scene with sufficient overlap between each other as a sample, where one frame is randomly chosen as the condition image (we set $N=4$ in all experiments). In training, we use the ground truth depth map rendered from the scene as the depth condition.



\noindent \textbf{Metrics.}
We evaluated the results of multi-view generation on the validation set of our dataset, consisting of 900 samples from 400 scenes.
We use the Inception Score (IS)~\cite{salimans2016improved} to assess visual quality and evaluate the novel-view synthesis consistency using PSNR and SSIM by comparing the generated $N-1$ views with the ground truth.

Additionally, we compared our method with other 3D scene generation approaches by analyzing the rendered images that depict the 3D scenes.
We conducted a user study to rate the perceptual quality (PQ), 3D structure consistency (3DC), and layout plausibility (LP) of generated views on a scale from 1 to 5.

\begin{table}[t]
\small
\centering
\vspace{-2mm}
\caption{Experimental results for fusion module.}
\vspace{-2mm}
\begin{tabular}{c|c|cc}
\toprule
Method  & IS$\uparrow$ & PSNR$\uparrow$  & SSIM$\uparrow$  \\
\hline
\hline
MVDiffusion~\cite{Tang2023mvdiffusion}        & 4.035   & 20.45        & 0.7247  \\
\hline
CAA      & 4.113   & 20.72       & 0.7435  \\
 3D self-attention        & \textbf{4.232}   &     22.06        & 0.7979 \\
Plain epipolar & 4.129 & 22.12 & 0.8077 \\
 LA-epipolar        & 4.170     & \textbf{22.66}        & \textbf{0.8154}   \\
\bottomrule
\end{tabular}
\vspace{-2mm}

\label{table:ablation_mv_fusion}
\end{table}

\begin{table}[t]
\small
\centering
\vspace{-1mm}
\caption{Ablation for model condtion priors.}
\vspace{-2mm}
\begin{tabular}{cccc|cc}
\toprule
single. & multi. & spatial & aligned dep. & PSNR$\uparrow$  & SSIM$\uparrow$  \\
\hline
\hline
\checkmark    &        &                      &      \checkmark       & 21.78        & 0.7863  \\
    &        \checkmark          &                  &     \checkmark      & 22.24        & 0.8044 \\
   &      \checkmark            & \checkmark           &            & 20.95        & 0.7579   \\
    & \checkmark       & \checkmark           &   \checkmark       & \textbf{22.66}        &  \textbf{0.8154} \\
\bottomrule
\end{tabular}
\label{table:ablation_layout_prior}
\vspace{-2mm}
\end{table}

\begin{table}[t]
\small
\centering
\vspace{-1mm}
\caption{User study results for multi-view scene generation. PQ: perception quality, 3DC: 3D structure consistency, LP: layout plausibility.}
\vspace{-2mm}
\begin{tabular}{l|ccc}
\toprule
Method  & PQ$\uparrow$ & 3DC$\uparrow$ & LP$\uparrow$  \\
\hline
\hline
Text2room~\cite{hollein2023text2room}        & 1.99  & 2.84  & 2.84  \\
LucidDreamer~\cite{chung2023luciddreamer}       & 1.93  & 2.72  & 2.78  \\
Set-the-Scene~\cite{cohen2023set}   & 1.53  & 2.79  & 2.59  \\
MVRoom (Ours)      & \textbf{4.59}  & \textbf{4.37}  & \textbf{4.21}  \\
\bottomrule
\end{tabular}

\label{table:scene_gen}
\vspace{-2mm}
\end{table}

\subsection{Analysis Experiments}

\begin{figure*}[h]
    \centering
    \includegraphics[width=0.92\linewidth]{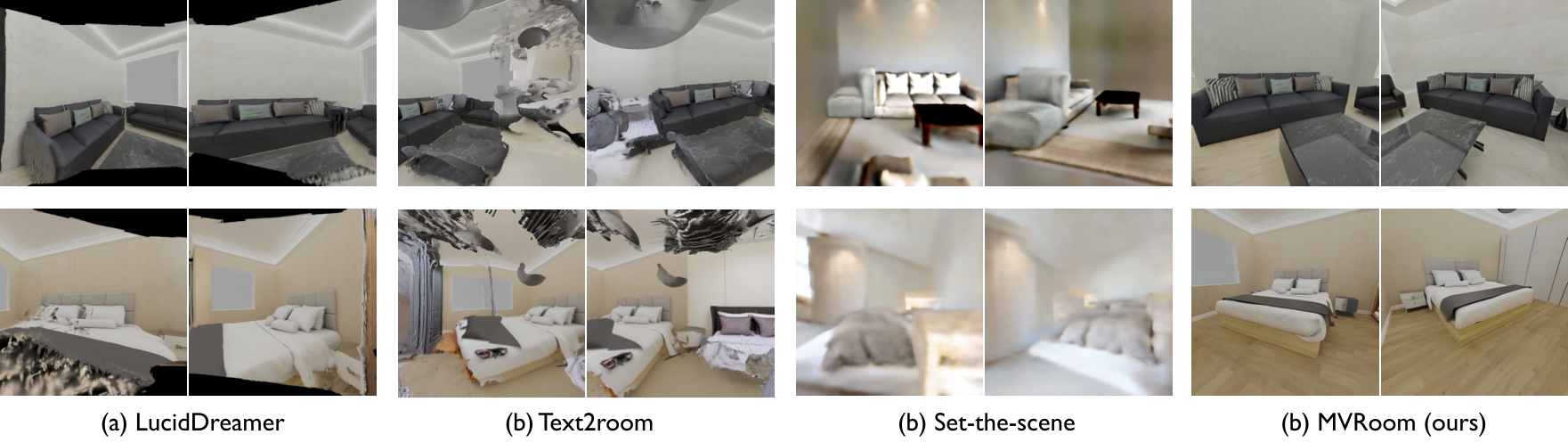}
    \vspace{-3mm}
    \caption{Qualitative results for scene generation methods. For LucidDreamer~\cite{chung2023luciddreamer} and Text2room~\cite{hollein2023text2room}, we use the same perspective view image as in our method and a text descrpition to generate a complete 3D scene. For Set-the-scene~\cite{cohen2023set}, we provide the scene layout along with a text description as input. The stable-diffusion-2-inpainting model used in baselines is fine-tuned on our dataset for fair comparison.}
    \vspace{-3mm}
    \label{fig:fig6_result2}
\end{figure*}

\begin{figure*}[h]
    \centering
    \includegraphics[width=1.0\linewidth]{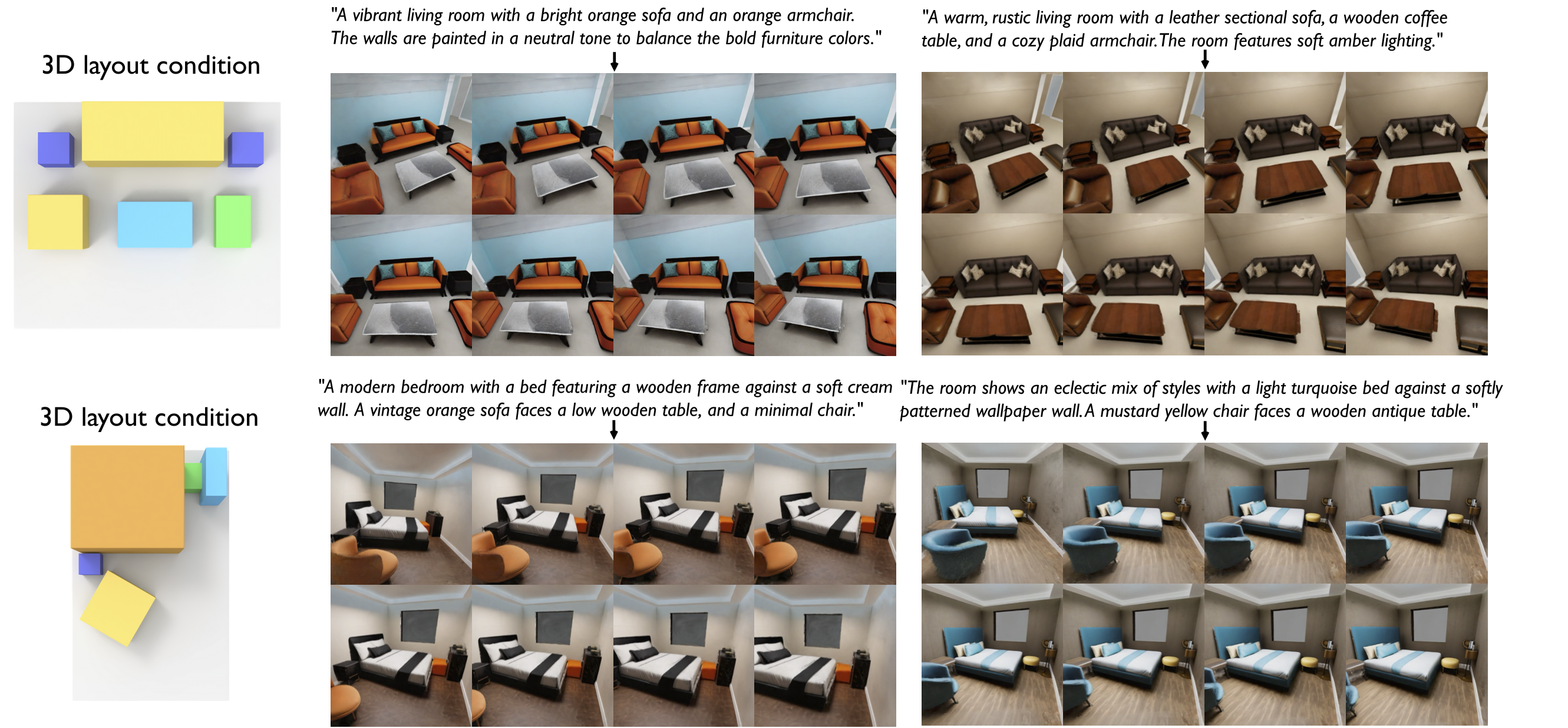}
    \vspace{-3.2mm}
    \caption{Text-to-scene generation. We show the 3D-GS rendering results with the corresponding 3D layout and text input.}
    \vspace{-3.5mm}
    \label{fig:fig7_text_to_scene}
\end{figure*}


\noindent \textbf{Multi-view fusion module design.}
Table~\ref{table:ablation_mv_fusion} compares MVRoom with MVDiffusion and other feature fusion modules used in the multi-view diffusion models. For MVDiffusion, we adopt its original implementation and incorporate our full layout adapters. In lines 2-5, we use correspondence-aware attention (CAA), 3D self-attention and plain epipolar attention to replace our layout-aware epipolar attention module. Specifically, CAA, from \cite{Tang2023mvdiffusion}, employs dense correspondence across views to calculate cross-attention, and the 3D self-attention module, which is widely used for multi-view generation~\cite{shi2023mvdream}, concatenates and inflates the 3D feature maps from all views to adapt to the original 2D self-attention.  For the plain epipolar attention module, we remove layout mask in the epipolar line calculation in Eq.~\ref{eq:epipolar}.

Our method significantly improves generation quality and multi-view consistency over MVDiffusion, with a dramatic 12.5\% increase in SSIM. Additionally, our proposed LA-epipolar attention module achieves notable consistency gains over all CAA, 3D self-attention, and plain epipolar attention designs, underscoring its efficacy in integrating camera parameters and 3D scene layout in feature aggregation.
Please refer to the supp. materials for more qualitative results.

\noindent \textbf{3D scene priors guidance.}
Table~\ref{table:ablation_layout_prior} compares the impact of different 3D layout condition signals on the consistency of multi-view generation. `single.' indicates using a single-layer semantics and depth condition ($m=1$), while `multi.' refers to a three-layer semantics and depth condition ($m=3$). Results show that multi-layer conditions effectively provide more comprehensive 3D scene priors, as they avoid the occlusion issues of a bounding box-based layout from the perspective view.
`spatial' denotes the use of layout spatial embedding. This embedding provides explicit pose and position information for object bounding boxes across all views, leading to improved consistency in generation.
`aligned dep.' refers to the use of the predicted depth for warping the conditioning image in Eq.~(\ref{eq:warp_img}). In line 3, we substitute the aligned depth with a coarse depth directly derived from layout bounding boxes. We see that using aligned depth as a geometry prior significantly enhances generation quality. Notably, even with only the layout's coarse depth, our method outperforms the full model setup with the CAA module in Table~\ref{table:ablation_mv_fusion}, highlighting the effectiveness of our LA-epipolar attention module.

\subsection{Scene Generation Comparison}

\noindent \textbf{Baselines.}
We compare our approach with open-source scene generation methods Set-the-Scene~\cite{cohen2023set}, Text2Room~\cite{hollein2023text2room}, and LucidDreamer~\cite{chung2023luciddreamer}. Set-the-Scene uses scene layout and text description as input, while Text2Room and LucidDreamer cannot take layout control as input. For evaluation, we selected scenes from the validation set of our dataset as the initial condition input for scene-level multi-view images generation. For Text2Room and LucidDreamer, we generated scenes using a perspective view image along with a textual prompt that describes the scene. 
We fine-tune the Stable Diffusion models used in the baselines on our dataset to ensure fair comparisons.
For each baseline, we rendered images from multiple views of the generated 3D representation for comparison.

\noindent \textbf{Results.}
We present the results of the user study in Table~\ref{table:scene_gen} and qualitative comparisons in Figure~\ref{fig:fig6_result2}. Human evaluation indicates that our method significantly outperforms all baseline methods in terms of perceptual quality, multi-view consistency, and layout plausibility of the generated results. 
We show in Figure~\ref{fig:fig1_overview} the generation results of our recursive scene generation framework.  
We can optimize a high-quality 3D Gaussian splatting (3D-GS) representation~\cite{kerbl20233d} with the consistent perspective images generated by MVRoom, allowing for high-fidelity and realistic image rendering from arbitrary viewpoints.
Furthermore, we demonstrate the text-to-scene generation results in Figure~\ref{fig:fig7_text_to_scene}. 
Given a text description as input, we leverage the trained adapter using $\mathbf{P}^i_{\textup{sem}}$ and $\mathbf{P}^i_{\textup{depth}}$ as conditions to generate an initial scene image from the first viewpoint. Results show that our text-to-scene framework produces style-diverse, high-quality 3D scenes that closely adhere to the provided 3D layout, enabling strong controllability.




\section{Conclusion}
\label{sec:conclusion}



We present MVRoom, a two-stage multi-view NVS model for controllable indoor scene generation, together with a novel layout-guided recursive scene generation framework supporting text-to-scene generation.
MVRoom features two core innovations: (1) a novel hybrid image-based conditioning mechanism that leverages 3D layouts and text-derived views for multi-view generation, and (2) a multi-view diffusion model with layout-aware epipolar attention. 
This design enables controllable, layout-guided generation across scenes of varying complexity while improving spatial alignment and cross-view consistency. 
Experimental results validate the effectiveness of our proposed method and several designed components. These findings underscore MVRoom's potential for immersive content creation, with applications in AR, VR, and virtual environment development.

{
    \small
    \bibliographystyle{ieeenat_fullname}
    \bibliography{main}
}



\section{Method}

\subsection{Depth Alignment}
\label{sec:depth_alignment}

Here we provide more details about the depth prediction and alignment process $\mathbf{\hat{D}}_{\textup{cond}} = g(\mathbf{X}_{\textup{cond}}, \mathcal{L})$ for condition image warping (Eq. (3)). 

Given a condition image $\textbf{X}$, we use a pre-trained depth estimation model~\cite{yang2024depth} to get a relative depth $\textbf{D} = f_d (\textbf{X})$ as initialization. Then we align the predicted depth $\textbf{D}$ according to the 3D layout. In section 3.1, we have the $m$-layer depth conditon $\mathbf{P}^i_{\textup{depth}} = ( \textbf{D}_1^l, \cdots, \textbf{D}_m^l )$. We can derive $\textbf{M}_{bg}$ and $\textbf{M}_{fg}$, which are the background mask and foreground mask, from $\textbf{D}_2^l$. We have $\textbf{M}_{fg} = (\textbf{D}_2^l > 0)$ and $\textbf{M}_{bg} = (\textbf{D}_2^l = 0)$, since during the rendering of $\mathbf{P}^i_{\textup{depth}}$, if a ray intersects with the background, all subsequent values are ignored and set as zero.

We first align the relative depth $\textbf{D}$ according to the background part of the layout prior depth. Following~\cite{hollein2023text2room}, we optimize the scale and offset parameters $s$ and $\delta$ by minimizing the least square error in the background regions

\begin{equation*}
    \min_{s, \delta} 
    \left\| 
     \left( \frac{s}{\hat{\textbf{D}}_p} + \delta - \frac{1}{\textbf{D}_1^l} \right)  \odot \textbf{M}_{bg} 
    \right\|^2
\end{equation*}

Then we can get an aligned depth map $\textbf{D}^{\prime}$ by $\frac{1}{\textbf{D}^{\prime}} = \frac{s}{\hat{\textbf{D}}_p} + \delta$. Additionally, since the the layout bounding box constrains the depth range for the foreground part of the predicted depth map, we further clip the aligned depth using the upper bound and lower bound of the depths based on the 3D layout. The lower bound is $\textbf{D}_1^l$, and the upper bound $\textbf{D}_{bg}^l$ is the last non-zero values in $\mathbf{P}^i_{\textup{depth}}$, which are the depth for background elements. Finally, we have $\mathbf{\hat{D}} = f_{clip} (\textbf{D}^{\prime}, \textbf{D}_1^l, \textbf{D}_{bg}^l, \textbf{M}_{fg})$, where $f_{clip}$ is a function that clip the foreground part values of the depth map.

\subsection{Camera Trajectory Generation}
\label{sec:view_selection_supp}




Here, we describe our approach for generating continuous camera trajectories in the scene to sample the perspective view poses for MVRoom training and generation. It is critical to select appropriate camera poses such that the visual content of the scene is extensively covered, with overlapping between adjacent views to ensure consistency during generation.

\begin{figure}[t]
    \centering
    \includegraphics[width=0.95\linewidth]{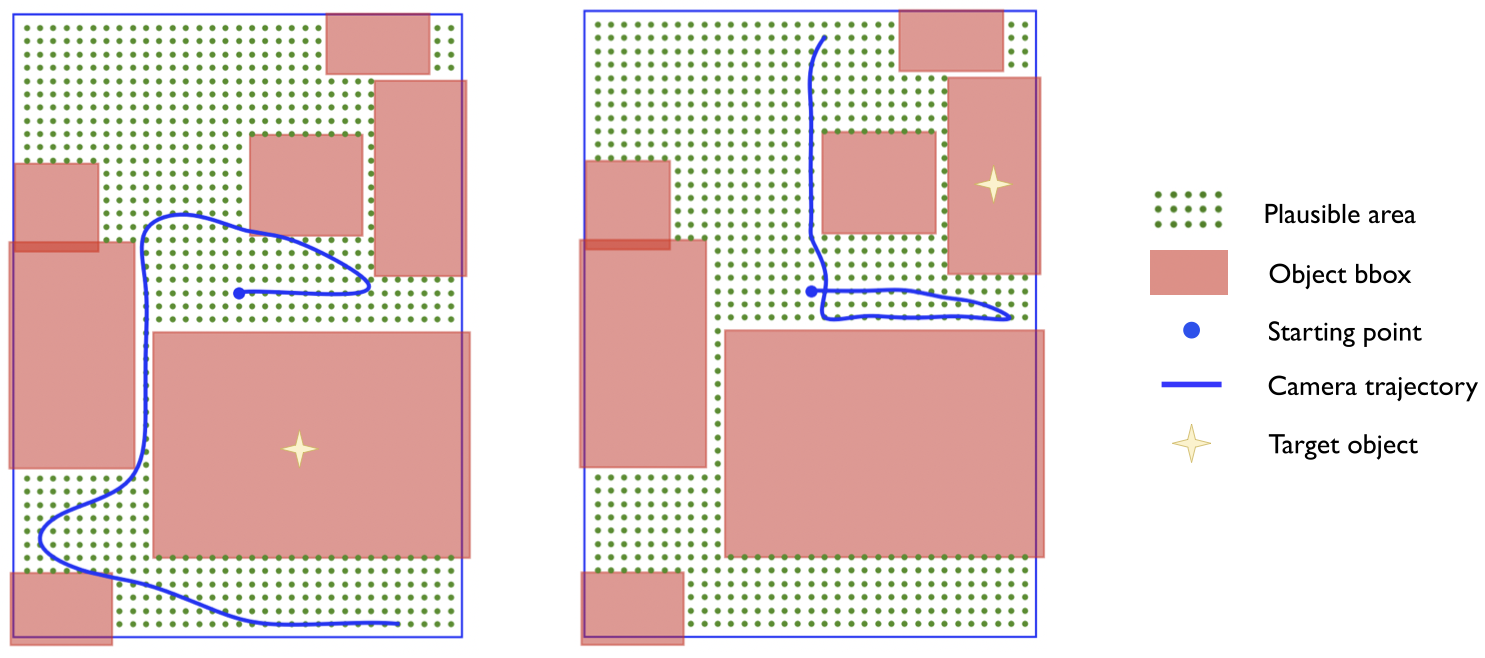}
    \caption{Illustration of two generated trajectories examples for a given scene with 3D layout. We first identify the plausible 2D area within the scene (indicated by the green dots in the figure). For a trajectory, we specify a target object and a starting point. An endpoint of the trajectory is chosen to be as far as possible from the starting point. The generated trajectory should 1) remain within the plausible area. 2) maintain an appropriate distance from the target object. }
    \label{fig:supp_fig_traj}
\end{figure}

To achieve this, we propose a camera trajectory generation method based on the 3D scene layout.
Each trajectory begins at an arbitrary point in the room and targets a specific object in the layout. The goal is to generate a continuous trajectory of camera poses that comprehensively captures the appearance of the target object.
First, the trajectory positions are sampled to maintain a reasonable distance from the target object. Also, to ensure physical plausibility, camera positions are constrained to unoccupied space within the room's 3D layout.
Therefore, we apply an A-star algorithm with these conditions to obtain feasible camera trajectory positions. Figure~\ref{fig:supp_fig_traj} illustrates an example of generated trajectories in an indoor scene given the 3D layout.
Once the 2D trajectory is obtained, we discretely sample camera poses along it. The sampling is based on either distance or angular changes in the camera’s position and orientation (e.g. using a 0.3m displacement interval).
Camera heights are randomly sampled from a range of (0.8m, 2.5m), and camera poses are oriented toward the center of the target object, while ensuring smooth transition and rotation.

\subsection{Recursive Scene Generation Details}

Algorithm 1 in the main paper presents the pseudocode of our proposed recursive scene generation method. Given a coarse 3D layout $\mathcal{L}$ and a text prompt describing the scene, our approach recursively generates dense scene images that cover the entire indoor space and reconstructs a complete scene representation.

\noindent \textbf{Initial Image}
The scene generation begins with an initial image $\mathbf{X}_0$, where its view $\mathbf{p}_0$ can be arbitrarily selected within the scene, to initialize the overall style and appearance. The initial image $\mathbf{X}_0$ can either be a rendered image or one generated from a text prompt based on the 3D layout.
To enable text-guided generation, we build upon a pre-trained text-to-image model and train separate adapters conditioned on the proposed multi-layer depth $\mathbf{P}_{\textup{depth}}$ and semantics $\mathbf{P}_{\textup{sem}}$. $\mathbf{P}_{\textup{depth}}$ and $\mathbf{P}_{\textup{sem}}$ can be derived from the layout and pose (Eq. (1)). In Algorithm 1, we refer to this layout-guided text-to-image generation model as $SD\_adapter(\cdot)$.

\noindent \textbf{Dense Views Generation}
Before starting the generation, we first generate the set of dense views within the scene.
Selecting appropriate camera views within a complex 3D scene for multi-view generation is a highly non-trivial task. These camera views must comprehensively cover the scene’s visual content. At the same time, it is crucial to ensure sufficient overlap between adjacent views to support consistent and efficient generation.
To achieve this, we adopt the camera trajectory generation method described in Sec 6.2. For each of the $N_o$ objects in the scene, we generate two trajectories from the initial view $\mathbf{p}_0$ at different heights. Camera positions along these trajectories are sampled at intervals of 0.4 meters.

After obtaining all these views, we apply a filtering step to remove redundant viewpoints. If two camera views are found to be too close, specifically, if both the positional distance and angular difference are below the thresholds (0.4 meters and 4 degrees in our implementation), one of them is discarded.
The remaining $N_p$ camera poses collected from these $2*N_o$ trajectories are used for the subsequent scene generation process.

\noindent \textbf{Generated Views Selection}
During the recursive generation process, we need to select appropriate viewpoints for novel view synthesis at each iteration.
We follow two rules. First, the selected views should have substantial overlap with the generated parts of the scene to enhance consistency. Second, we prioritize choosing consecutive views along the same trajectory to better align with the view sampling strategy used during training.

In implementation, we maintain the ordered list of ungenerated view indices for each trajectory. At each generation step, we compare the first ungenerated view on each trajectory and compute its overlap with the current global point cloud, and then select the view with the highest overlap. The overlap ratio is estimated by rendering the global point cloud from each candidate view.
Once the best candidate view is determined, denoted as $\mathbf{p}_{k_1}$, we select the $N$ consecutive views $\mathbf{p}_{k_1}, \cdots, \mathbf{p}_{k_N}$ along that trajectory for subsequent MVRoom generation. If no views can be selected using this method, the generation iteration is terminated.

\noindent \textbf{Global Point Cloud Update}
During generation across different batches, the visual and geometric consistency of the scene can not be guaranteed. 
To mitigate the inconsistency issue, we assess the geometric consistency of each generated result by comparing the predicted depth from the novel view image with the depth rendered from the global point cloud.

For a generated image $\mathbf{X}_{k_i}$, we obtain a predicted depth map $\mathbf{\hat{D}}_{k_i}$ using the depth estimation method $g(\cdot)$ (see Sec.~\ref{sec:depth_alignment}). Simultaneously, we render a depth map $\mathbf{D}_{k_i}$ from the current global point cloud using the pose $\mathbf{p}_{k_i}$ associated with $\mathbf{X}_{k_i}$. Since $\mathbf{p}_{k_i}$ is a novel view within the scene, the rendered depth map $\mathbf{D}_{k_i}$ usually contains holes. We denote the valid region of $\mathbf{D}_{k_i}$ as a binary mask $\mathbf{M}_{k_i}$.
If the newly generated content $\mathbf{X}_{k_i}$ is geometrically consistent with the existing scene, the predicted depth $\mathbf{\hat{D}}_{k_i}$ should closely match the rendered depth $\mathbf{D}_{k_i}$ within the valid mask $\mathbf{M}_{k_i}$. 
Therefore, we can formulate the geometric inconsistency as $\left\|  (\mathbf{D}_{k_i} - \mathbf{\hat{D}}_{k_i}) \odot \mathbf{M}_{k_i}  \right\|^1$.
When updating the global point cloud, we only integrate the newly predicted point cloud if the inconsistency metric is below a predefined threshold (the threshold is set to 0.02m in our implementation).

\section{Implementation Details}
\noindent \textbf{Training.}
We use the pre-trained weights of Stable Diffusion v2-1~\cite{von-platen-etal-2022-diffusers} as the model initialization and keep its parameters frozen during training.
For layout-conditioned text-to-image generation, we train the adapters on 8 H20 GPUs for 10k steps, with a batch size of 8 and a learning rate of 1e-4. 
Then we train the multi-view diffusion model on 24 H20 GPUs for 60k steps, with a batch size of 1 and a learning rate of 1e-4, and the inference and test are conducted on the same machine. For image-conditioned multi-view generation tasks, the number of views $N$ is set to 4, with a resolution of $512 \times 512$ for both training and validation.

\noindent \textbf{Computaton Cost.} In implementation, our method computes the image-based layout conditions (in Eq. (1) and Eq. (2)) and epipolar masks (in Eq. (5)) in advance, which takes an extra time of 0.3s and 2.2s in average, respectively, for each multi-view generation on the same device as used during training.

\noindent \textbf{User Study Details.} 
In our user study (Table 3), we asked 20 participants to assess 3D scenes generated by different methods. Participants rated each scene individually on a scale of 1 to 5 for three specific aspects: perceptual quality (PQ), 3D structure consistency (3DC), and layout plausibility (LP). The questions for each aspect were: (1) Perceptual Quality (PQ): Please rate the clarity, detail, and overall visual quality of the generated images. (2) 3D Structure Consistency (3DC): Please rate whether the same scene appears consistent and undistorted from different viewpoints. (3) Layout Plausibility (LP): Please rate how well the arrangement of objects and the background resemble a realistic indoor room setting.

\section{Dataset Preparation}

Our training and validation datasets are constructed using the 3D-FRONT dataset, which offers high-quality 3D indoor scenes with professionally designed layouts and textured object models.
To prepare the data, we first segment the 3D-FRONT scenes into individual rooms and filter out those with unreasonable layouts, resulting in 6,287 valid rooms (5,887 for training and 400 for validation). 
Using the camera trajectory generation method described in Sec.~\ref{sec:view_selection_supp}, we generate multiple camera trajectories for each room. Camera positions along these trajectories are sampled at 0.2m intervals, and RGB images are rendered for each pose using Blender.

To extract required conditions, such as depth and semantics, we project the 3D layout's bounding boxes onto the images based on the camera poses. For rendering multi-layer conditions (as outlined in Eq. (1)), we separately render the object bounding boxes and background layout components, then combine them using depth relationships.
Additionally, we perform filtering to remove low-quality data based on depth and semantics. Samples with depth values below 0.5m or a foreground semantics proportion under 0.2 are excluded. After the filtering process, the dataset retains approximately 4 million images for training and validation.


\section{Visualization results}


We show more multi-view generation results of MVRoom in Figure~\ref{fig:fig9_mv} and scene generation results in Figure~\ref{fig:fig8_recon}.

\begin{figure}[t]
    \centering
    \includegraphics[width=0.95\linewidth]{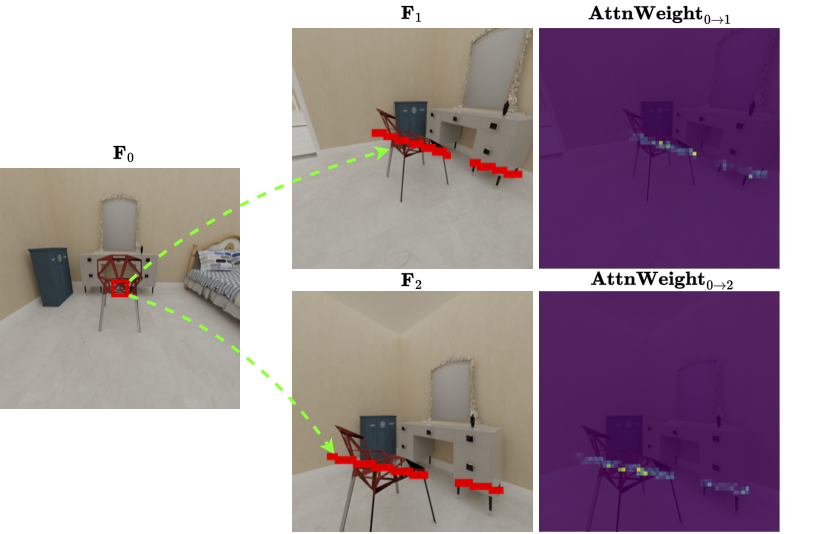}
    \caption{Visualization of the Layout-aware Epipolar attention map. We use epipolar geometry and the 3D layout to explicitly identify regions of potential visual content overlap between views. During generation, cross-attention mechanisms are applied to maintain consistency across multiple views.}
    \label{fig:supp_fig_attn_weight}
\end{figure}


\begin{figure*}[h]
    \centering
    \includegraphics[width=0.95\linewidth]{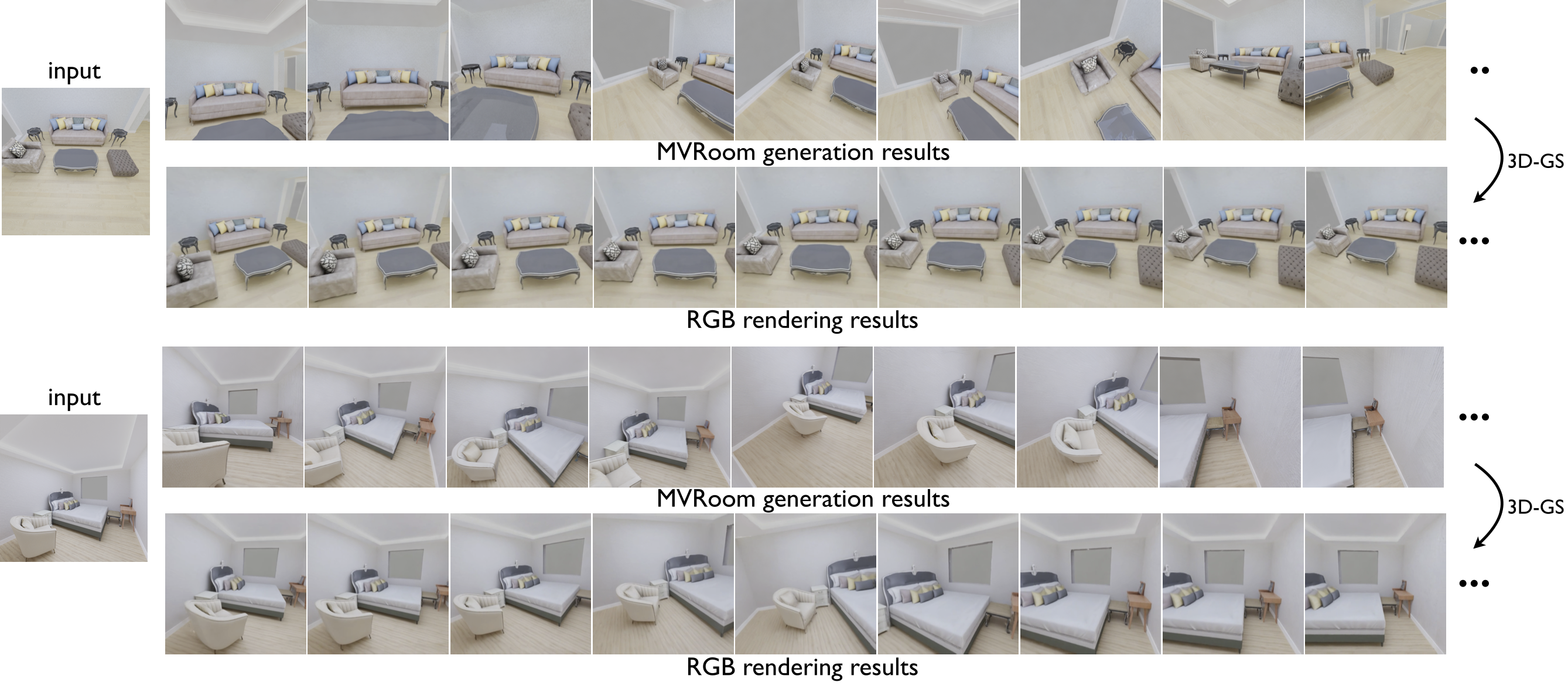}
    \caption{Scene generation results. We use a 3D scene layout and a corresponding single image as input. We show a collection of different view results generated recursively by MVRoom, as well as RGB rendering results from the 3D Gaussian splatting optimization.}
    \label{fig:fig8_recon}
\end{figure*}

\begin{figure*}[h]
    \centering
    \includegraphics[width=0.85\linewidth]{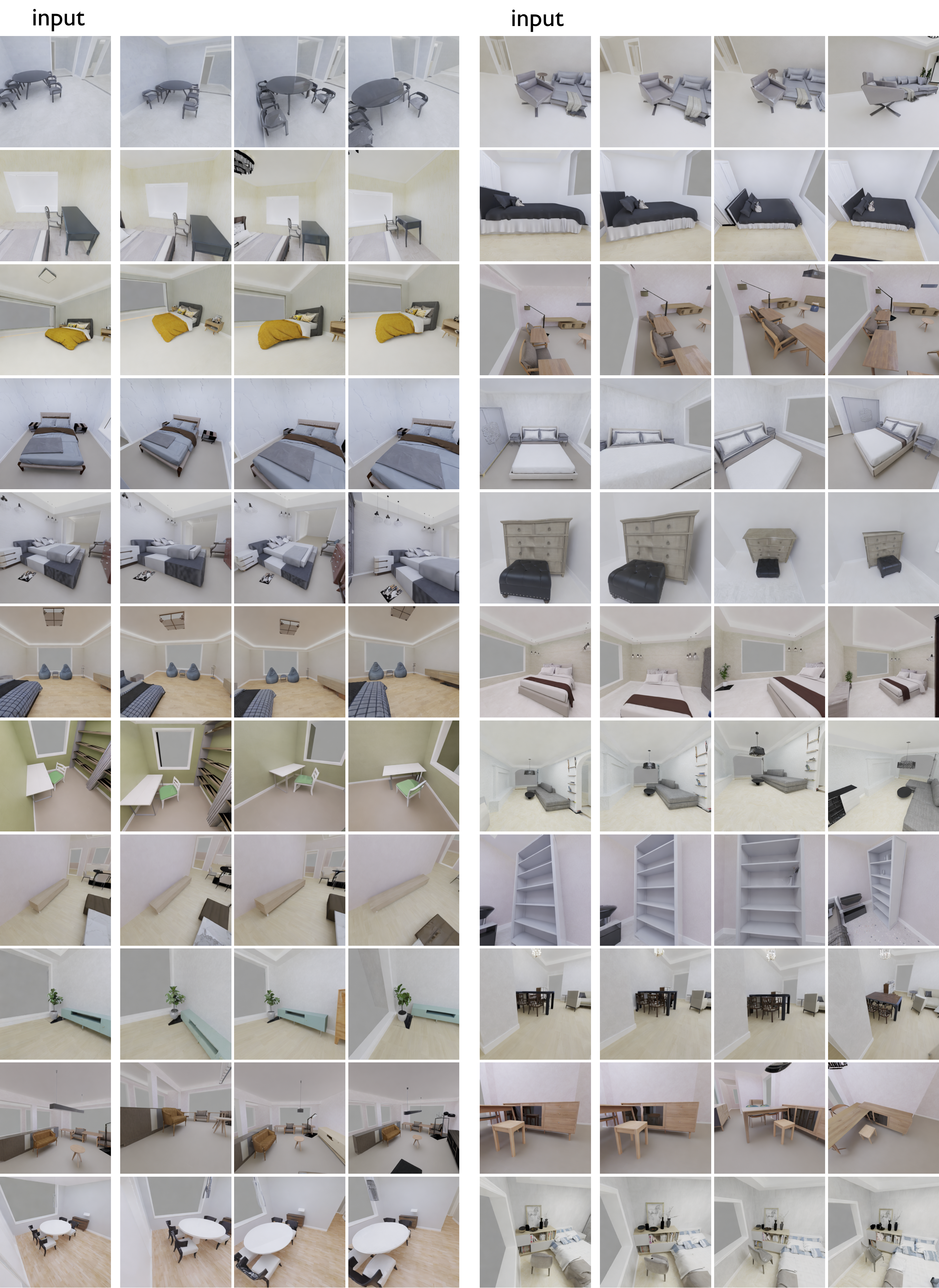}
    \caption{Additional qualitative results on 3D-FRONT~\cite{3dfront} validation set}
    \label{fig:fig9_mv}
\end{figure*}






\section{Limitation and Future Work}

While our method demonstrates promising performances in layout-controlled novel view synthesis and text-guided scene generation, it still has several limitations.
First, the current framework only supports generating scenes from rendered images or text-generated images. A more practical application would be extending the scene generation capability to real-world or more complex scene images while maintaining layout consistency. 
Second, our model's capabilities are restrained by the Stable Diffusion foundation model. Currently, video foundation models exhibit stronger capabilities in generating consistent views. A future direction involves effectively integrating 3D layout priors to guide the generation process of video foundation models.

\end{document}